\documentclass[11pt,a4paper]{article}
\usepackage[hyperref]{eacl2021}
\usepackage{times}
\usepackage{latexsym}
\usepackage{amsmath}
\usepackage{graphicx}
\usepackage{footnote}
\usepackage{booktabs}
\usepackage{xspace}
\usepackage{bm}
\usepackage{todonotes}

\newcommand{\ignore}[1]{}

\usepackage{microtype}

\aclfinalcopy 

\newcommand{\Sref}[1]{\S\ref{#1}}

\newcommand{\Fref}[1]{Figure~\ref{#1}}

\newcommand{\ourmodel}{{StructSum}\xspace}

\title{\ourmodel: Summarization via Structured Representations}

\author{Vidhisha Balachandran \qquad Artidoro Pagnoni
\qquad Jay Yoon Lee \\
\qquad \textbf{Dheeraj Rajagopal} \qquad \textbf{Jaime Carbonell} \qquad \textbf{Yulia Tsvetkov}\\
School of Computer Science \\
Carnegie Mellon University, Pittsburgh, USA \\
\texttt{\{vbalacha,apagnoni,jaylee,dheeraj,jgc,ytsvetko\}@cs.cmu.edu}
}

\date{}

\begin{document}
\maketitle
\begin{abstract}
Abstractive text summarization aims at compressing the information of a long source document into a rephrased, condensed summary. Despite advances in modeling techniques, abstractive summarization models still suffer from several key challenges: (i) \emph{layout bias}: they overfit to the style of training corpora; (ii) \emph{limited abstractiveness}: they are optimized to copying n-grams from the source rather than generating novel abstractive summaries; (iii) \emph{lack of transparency}: they are not interpretable.
In this work, we propose a framework based on document-level structure induction for summarization to address these challenges. To this end, we propose incorporating \emph{latent and explicit dependencies across sentences in the source document} into end-to-end single-document summarization models. Our framework complements standard encoder-decoder summarization models by augmenting them with rich structure-aware document representations based on implicitly learned (latent) structures and externally-derived linguistic (explicit) structures. We show that our summarization framework, trained on the CNN/DM dataset, improves the coverage of content in the source documents, generates more abstractive summaries by generating more novel n-grams, and incorporates interpretable sentence-level structures, while performing on par with standard baselines.\footnote{Code and data available at: \url{https://github.com/vidhishanair/structured\_summarizer} } 
\end{abstract}

\section{Introduction}
Text summarization aims at identifying important information in long source documents and expressing it in human readable summaries. Two prominent methods of generating summaries are \emph{extractive} \cite{dorr2003hedge, nallapati2017summarunner}, where important sentences in the source article are selected to form a summary,  and \emph{abstractive} \cite{rush-2015-sum, see-etal-2017-get}, where the model restructures and rephrases essential content into  a paraphrased summary. 

State of the art approaches to abstractive summarization employ neural encoder-decoder methods that encode the source document as a sequence of tokens producing latent document representations and decode the summary conditioned on the representations. Recent studies suggest that these models suffer from several key challenges. First, since standard training datasets are derived from news articles,  model outputs are strongly affected by the layout bias of the articles, with models relying on the leading sentences of source documents \cite{kryscinski-etal-2019-neural, kedzie-etal-2018-content}. Second, although they aim to generate paraphrased summaries, abstractive summarization systems often copy long sequences from the source, causing their outputs to resemble extractive summaries \cite{lin2019abstractive, Gehrmann2018BottomUpAS}. Finally,  current methods do not lend themselves easily to interpretation via intermediate structures \cite{lin2019abstractive},  which could be useful for identifying major bottlenecks in summarization models.

To address these challenges, we introduce \emph{\ourmodel}: a framework that incorporates 
structured document representations into summarization models. \ourmodel complements a standard encoder-decoder architecture with two novel components: (1) a \emph{latent-structure attention} module that adapts structured representations \citep{Kim2017StructuredAN, Liu2017LearningST} for the summarization task, and (2) an \emph{explicit-structure attention} module that incorporates an external linguistic structure (e.g., coreference links). The two complementary components are incorporated and learned jointly with the encoder and decoder, as shown in \Fref{fig:eg2}.

Encoders with induced latent structures have been shown to benefit several tasks including document classification, natural language inference \citep{Liu2017LearningST, cheng-etal-2016-long}, and machine translation \citep{Kim2017StructuredAN}. Our latent structure attention module builds upon \citet{Liu2017LearningST} to model the dependencies between sentences in a document. It uses a variant of Kirchhoff's matrix-tree theorem \citep{tutte1984graph} to model such dependencies as non-projective tree structures(\Sref{sec:LS}). The explicit attention module is linguistically-motivated and aims to incorporate inter-sentence links from externally annotated document structures. We incorporate a coreference based dependency graph across sentences, which is then combined with the output of the latent structure attention module to produce a hybrid structure-aware sentence representation (\Sref{sec:ES}).

We test our framework using the CNN/DM dataset \citep{hermann2015teaching} and show in \Sref{sec:rouge} that it outperforms the base pointer-generator model \cite{see-etal-2017-get} by up to 1.1 ROUGE-L. We find that the latent and explicit structures are complementary, both contributing to the final performance improvement. Our modules are also orthogonal to the choice of an underlying encoder-decoder architecture, rendering them flexible to be incorporated into other advanced models.

Quantitative and qualitative analyses of summaries generated by \ourmodel and baselines (\Sref{sec:results}), reveal that structure-aware summarization mitigates the news corpora layout bias by improving the coverage of source document sentences. Additionally, \ourmodel reduces the bias of copying large sequences from the source, inherently making the summaries more abstractive by generating $\sim$15\% more novel n-grams than a competitive baseline. We also show examples of the learned interpretable sentence dependency structures, motivating further research for structure-aware modeling.

\section{\ourmodel Framework}
\label{sec:model}

Consider a source document $\bm{x}$ consisting of $n$ sentences $\{\bm{s}\}$ where each sentence $\bm{s}_i$ is composed of a sequence of words. Document summarization aims to map the source document to a target summary $\bm{y}$ of $m$ words $\{y\}$.
A typical neural abstractive summarization system is an attentional sequence-to-sequence model that encodes the input sequence $\bm{x}$ as a continuous sequence of tokens $\{w\}$ using a standard encoder \citep{hochreiter1997long, vaswani2017attention}. The encoder produces a set of hidden representations $\{\mathbf{h}\}$. A decoder maps the previously generated token  $y_{t-1}$ to a hidden state and computes a soft attention probability distribution $p(\mathbf{a}_t \mid \bm{x}, \bm{y}_{1:t-1})$ over encoder hidden states. A distribution  $p$ over the vocabulary is computed at every time step $t$ and the network is trained using the negative log likelihood loss:
$\text{loss}_t = - \mathrm{log}\:p(y_t) $. 

\begin{figure*}
    \centering
    \includegraphics[width=1.0\textwidth]{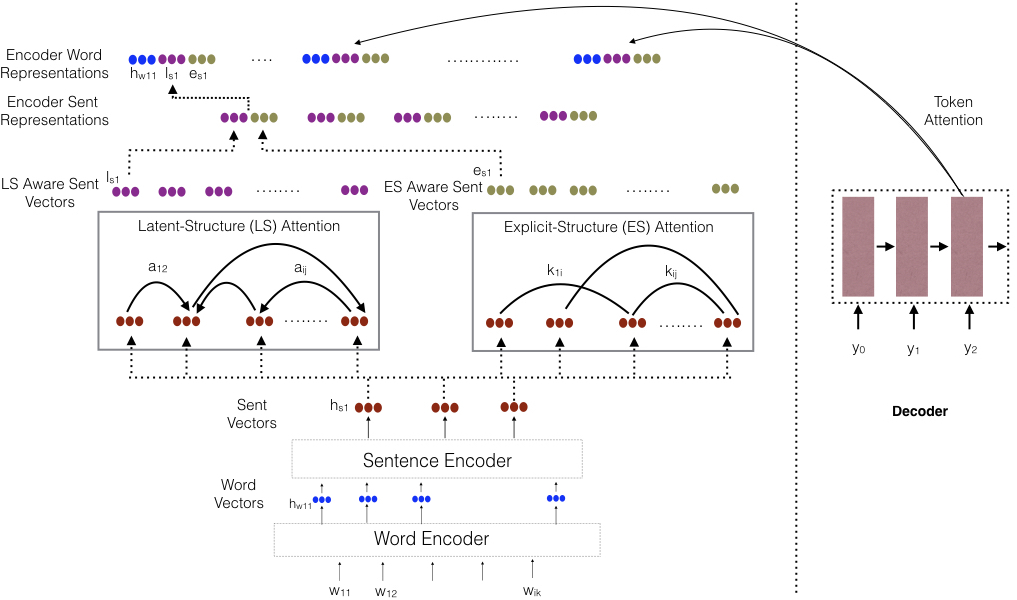}
    \caption{\ourmodel incorporates Latent Structure (LS) \Sref{sec:LS} and Explicit Structure (ES) \Sref{sec:ES} Attention to produce structure-aware representations. Here, \ourmodel augments the Pointer-Generator model, but the methodology that we proposed is general, and it can be applied to other encoder-decoder summarization systems}
    \label{fig:eg2}
\end{figure*}

\ourmodel modifies the above architecture as follows. 
We aggregate the token representations from the encoder to form sentence representations as in hierarchical encoders \cite{Yang2016HierarchicalAN}. We then use implicit- and explicit-structure attention modules to augment the sentence representations with sentence dependency information, leveraging both a learned latent structure and an external structure from other NLP modules. The attended vectors are then passed to the decoder, which produces the output abstractive summary.
In the rest of this section, we describe our framework architecture, shown in \Fref{fig:eg2}, in detail.

\subsection{Sentence Representations}
We consider an encoder which takes a sequence of words in a sentence $\bm{s}_i = \{w\}$ as input and produces contextual hidden representation for each word $\mathbf{h}_{w_{ik}}$, where $w_{ik}$ is the $k^{th}$ word of the $i^{th}$ sentence, $k=1:q$ and $q$ is the number of words in the sentence $\bm{s}_i$. The word hidden representations are max-pooled at the sentence level and passed through a sentence-encoder, which produces new hidden sentence representations for each sentence $\mathbf{h}_{\bm{s}_i}$. The sentence hidden representations are then passed as inputs to the latent and explicit structure attention modules.

\subsection{Latent Structure (LS) Attention}
\label{sec:LS}
We model the latent structure of a source document as a non-projective dependency tree of sentences and force a pairwise attention module to automatically induce this tree. We denote the marginal probability of a dependency edge as $a_{ij} = p(z_{ij}=1)$ where $z_{ij}$ is the latent variable representing the edge from sentence $i$ to sentence $j$. We parameterize the unnormalized pairwise scores between sentences with a neural network and use the Kirchoff's matrix tree theorem \cite{tutte1984graph} to compute the marginal probability of a dependency edge between any two sentences.

Specifically, we decompose the representation of a sentence $\bm{s}_i$ into a \emph{semantic} vector $\mathbf{g}_{\bm{s}_i}$ and \emph{structure} vector $\mathbf{d}_{\bm{s}_i}$ as $\mathbf{h}_{\bm{s}_i} = [\mathbf{g}_{\bm{s}_i}; \mathbf{d}_{\bm{s}_i}]$. Using the structure vectors $\mathbf{d}_{\bm{s}_i}, \mathbf{d}_{\bm{s}_j}$, we compute a score $f_{ij}$ between sentence pairs $(i,j)$ (where sentence $i$ is the parent node of sentence $j$) and a score $r_i$ (where the sentence $\bm{s}_i$ is the root node):
\vspace{-0.2cm}
\begin{align}
    f_{ij} = F_p(\mathbf{d}_{\bm{s}_i})^T W_a F_c(\mathbf{d}_{\bm{s}_j}) \text{ and } 
    r_i = F_r(\mathbf{d}_{\bm{s}_i}) \nonumber
\end{align}

where $F_p, F_c$ and $F_r$ are linear-projection functions that build representations for the parent, child and root nodes respectively, and $W_a$ is the weight for bilinear transformation. Here, $f_{ij}$ is the edge weight between nodes $(i,j)$ in a weighted adjacency graph $\mathbf{F}$ and is computed for all pairs of sentences. Using $f_{ij}$ and $r_i$, we compute normalized attention scores $a_{ij}$ and $a_{i}^r $ using a variant of Kirchhoff’s matrix-tree theorem where $a_{ij}$ is the marginal probability of a dependency edge between sentences $(i,j)$ and $a_{i}^r $ is the probability of sentence $i$ being the root.

Using these probabilistic attention weights and the semantic vectors $\{ \mathbf{g}_{\bm{s}}\}$, we compute the attended sentence representations as:
\vspace{-0.4cm}
\begin{align*}
    \mathbf{p}_{\bm{s}_i} &= \sum_{j=1}^{n} a_{ji} \mathbf{g}_{\bm{s}_j} + a_{i}^r \mathbf{g}_{\textit{root}} \\  
    \mathbf{c}_{\bm{s}_i} &= \sum_{j=1}^{n} a_{ij} \mathbf{g}_{\bm{s}_i}\\
    \mathbf{l}_{\bm{s}_i} &= \tanh(W_r[\mathbf{g}_{\bm{s}_i}, \mathbf{p}_{\bm{s}_i}, \mathbf{c}_{\bm{s}_i}])
\end{align*} 
\vspace{-0.05cm}
where $\mathbf{p}_{\bm{s}_i}$ is the context vector gathered from possible parents of sentence $i$, $\mathbf{c}_{\bm{s}_i}$ is the context vector gathered from possible children, and $\mathbf{g}_{root}$ is a special embedding for the root node. Here, the updated sentence representation $\textit{l}_{\bm{s}_i}$ incorporates the implicit structural information. 

\subsection{Explicit Structure (ES) Attention}
\label{sec:ES}
Following \citet{Durrett2016LearningBasedSS}, who showed that modeling coreference knowledge through anaphora constraints leads to improved clarity or grammaticality, we incorporate cross-sentence coreference links as the source of explicit structure. First, we use an off-the-shelf coreference parser\footnote{\url{https://github.com/huggingface/neuralcoref/}} to identify coreferring mentions. We then build a coreference based sentence graph by adding a link between sentences $(\bm{s}_i, \bm{s}_j)$, if they have any coreferring mentions. This graph is converted into a weighted graph by incorporating a weight on the edge between two sentences that is proportional to the number of unique coreferring mentions between them. We normalize these edge weights for every sentence, effectively building a weighted adjacency matrix $\mathbf{K}$ where $k_{ij}$ is given by:

\begin{align}
    k_{ij} &= P(z_{ij}=1) \\
    &= \frac{count(m_i \bigcap m_j)+\epsilon}{\sum_{v=1}^{n} count(m_i \bigcap m_v) }
\end{align}
where $m_i$ denotes the set of unique mentions in sentence $\bm{s}_i$, ($m_i$ $\bigcap$ $m_j$) denotes the set of co-referring mentions between the two sentences, and $z$ is a latent variable representing a link in the coreference sentence graph. $\epsilon = 5e^{-4}$ is a smoothing hyperparameter.

Given contextual sentence representations $\{\mathbf{h}_{\bm{s}}\}$ and our explicit coreference-based weighted adjacency matrix $\mathbf{K}$, we learn an explicit structure-aware representation as follows:
\begin{align*}
    \mathbf{u}_{\bm{s}_i} &= \tanh(F_u(\mathbf{h}_{\bm{s}_i})) \\
    \mathbf{t}_{\bm{s}_i} &= \sum_{j=1}^{p} k_{ij} \mathbf{u}_{\bm{s}_j} \\
     \mathbf{e}_{\bm{s}_i} &= \tanh(F_e(\mathbf{t}_{\bm{s}_i}))
\end{align*}
where $F_u$ and $F_e$ are linear projections and $\mathbf{e}_{\bm{s}_i}$ is an updated sentence representation which incorporates explicit structural information. \\

Finally, to combine the two structural representations, we concatenate the latent and explicit sentence vectors as: $\mathbf{h}_{\bm{s}_i} = [\mathbf{l}_{\bm{s}_i};\mathbf{e}_{\bm{s}_i}]$ to form encoder sentence representations of the source document.
To provide every token representation with the context of the entire document, the token representations are concatenated with their corresponding structure-aware sentence representation: $\mathbf{h}_{w_{ij}} = [\mathbf{h}_{w_{ij}};\mathbf{h}_{\bm{s}_i}]$ where $\bm{s}_i$ is the sentence to which the word $w_{ij}$ belongs. The resulting structure-aware token representations can be used to directly replace previous token representations as input to the decoder.

\section{Experiments}
\paragraph{Dataset:}
We evaluate our approach on the CNN/Daily Mail corpus\footnote{\url{https://cs.nyu.edu/~kcho/DMQA/}} \citep{hermann2015teaching, Nallapati2016AbstractiveTS} and use the same preprocessing steps as shown in \citet{see-etal-2017-get}. The CNN/DM has 287226/13368/11490 train/val/test samples respectively. The reference summaries have an average of 66 tokens ($\sigma = 26$) and 4.9 sentences. Differing from \citet{see-etal-2017-get}, we truncate source documents to 700 tokens instead of 400 in training and validation sets to model longer documents with more sentences. All our experiments were trained on Nvidia GTX Titan X GPUs.

\paragraph{Base Model:}
Although \ourmodel framework can be incorporated in any encoder-decoder framework with structure-aware representations, for our experiments we chose the pointer-generator model \citep{see-etal-2017-get} as the base model, due to its simplicity and ubiquitous usage as a neural abstractive summarization model across different domains \cite{Liu2019TopicAwarePN, krishna2020generating}. The word and sentence encoders are BiLSTM and the decoder is a BiLSTM with a pointer based copy mechanism.  We re-implement the base pointer-generator model and augment it with the \ourmodel modules described in \Sref{sec:model} and hence our model can be directly compared to it.
\paragraph{Baselines:}

In addition to the base model, we compare \ourmodel with the following baselines:\\
\citet{tan2017abstractive}: This is a graph-based attention model that is closest in spirit to the method we present in this work. A graph attention module is used to learn attention between sentences, but it cannot be easily used to induce interpretable document structures, since its attention scores are not constrained to learn structure. On top of latent and interpretable structured attention between sentences, \ourmodel introduces an explicit structure component to inject external document structure, which distinguishes it from  \citet{tan2017abstractive}. \\
\citet{Gehrmann2018BottomUpAS}: This work introduces a separate content selector which tags words and phrases to be copied. The DiffMask variant is an end-to-end variant like ours and hence is included in our baselines. We compare \ourmodel with the DiffMask experiment.\footnote{The best results from \citet{Gehrmann2018BottomUpAS} outperform DiffMask experiment, 
but they use inference-time hard masking which can be applied on ours.
Our baselines also exclude Reinforcement Learning (RL) based systems as they are not directly comparable, but our approach can be introduced in an encoder-decoder based RL system. Since we do not incorporate any pretraining, we do not compare with recent contextual representation based models \citep{Liu2019TextSW}.}

 \begin{savenotes}
\begin{table*}[t!]
\begin{center}
\begin{tabular}{@{}lllll@{}}
\hline \bf Model & \bf ROUGE 1 & \bf ROUGE 2 & \bf ROUGE L \\ \hline
Pointer-Generator \cite{see-etal-2017-get} & 36.44 & 15.66 & 33.42 \\
Pointer-Generator + Coverage \cite{see-etal-2017-get} &39.53 & \textbf{17.28} & 36.38\\
Graph Attention \cite{tan2017abstractive} & 38.10 & 13.90 & 34.00 \\
Pointer-Generator + DiffMask \cite{Gehrmann2018BottomUpAS} & 38.45 & 16.88 & 35.81\\
\midrule
Pointer-Generator (Re-Implementation) & 35.55 &	15.29 &	32.05\\
Pointer-Generator + Coverage (Re-Implementation) & 39.07 & 16.97 & 35.87\\
Latent-Structure (LS) Attention & 39.52 & 16.94 & 36.71 \\
Explicit-Structure (ES) Attention & \textbf{39.63} & 16.98 & 36.72 \\
 LS + ES Attention & 39.62 & 17.00 & \textbf{36.95} \\
\hline
\end{tabular}
\end{center}
\caption{\label{tab:summ-results}  Evaluation of summarization models on the CNN/DM dataset. Published abstractive summarization baseline scores are on top. The bottom section shows re-implementations of \citet{see-etal-2017-get}\footnote{ \url{https://github.com/atulkum/pointer_summarizer}} and \ourmodel results that incorporate latent and explicit document structure into the base models. \ourmodel's utility is on par with the base models, while introducing additional benefits of better abstractiveness and intrepretability shown~in~\Sref{sec:results}. } 
\end{table*}
\end{savenotes}

\paragraph{Hyperparameters:}
Our encoder uses 256 hidden states for both directions in the one-layer BiLSTM, and 512 for the single-layer decoder. We use the Adagrad optimizer \citep{duchi2011adaptive} with a learning rate of 0.15 and an initial accumulator value of 0.1. We do not use dropout and use gradient-clipping with a maximum norm of 2. We selected the best model using early stopping based on the ROUGE score on the validation dataset as our criteria. We also used the coverage penalty during inference as shown in \citet{Gehrmann2018BottomUpAS}. For decoding, we use beam-search with a beam width of 3. We did not observe significant improvements with higher beam widths.

\section{Evaluation}
\label{sec:results}
A standard ROUGE metric does not shed meaningful light into the quality of summaries across important dimensions. As a recall-based metric it is not suitable for assessing the abstractiveness of summarization; it is also agnostic to layout biases and does not facilitate intrepretability of model decisions. We thus adopt automatic metrics tailored to evaluating separately each of these aspects. We compare \ourmodel to our base model, the pointer-generator network with coverage \citep{see-etal-2017-get} and the reference.

\subsection{Automatic Metrics }
\label{sec:rouge}
We first conduct a standard comparison of generated summaries with reference summaries using ROUGE-1,2 and L \cite{lin2004rouge} F1\footnote{\url{https://pypi.org/project/pyrouge/}} metric. Table \ref{tab:summ-results} shows the results. We first observe that introducing the latent structures and explicit structures independently improves our performance on ROUGE-L. It suggests that modeling dependencies between sentences helps the model compose better long sequences compared to baselines. We see small improvements in ROUGE-1 and ROUGE-2, hinting that we retrieve similar content words as the baseline but compose them into better contiguous sequences. As both ES and LS independently get similar performance, the results show that LS attention induces good latent dependencies that make up for pure external coreference knowledge. 

Finally, our combined model which uses both Latent and Explicit structure performs the best with an improvement of \textbf{1.08 points} in ROUGE-L and \textbf{0.6 points} in ROUGE-1 over base pointer-generator model (statistically significant for 11490 samples at p=0.05 using Wilson Confidence Test). It shows that the latent and explicit information are complementary and a model can jointly leverage them to produce better summaries. Additionally, we find that structural inductive bias helps a model to converge faster. The combined LS+ES Attention model converges in 126K iterations in comparison to $\sim$230K iterations required for the pointer-generator network.

While ROUGE is a popular metric used for evaluating summarization models, it is limited to only evaluating n-gram overlap while ignoring semantic correctness. Hence, we compared our method with the baseline Pointer-Generator model using the BERTScore metric \cite{Zhang2020BERTScoreET}. We observe that our model improves BERTScore by ~9 points (12.3 for Pointer-Generator v/s 21.7 for StructSum) showing that our model is able to generate semantically correct content.

\subsection{Abstractiveness}
\label{sec:abs}
Despite being an abstractive model, the pointer-generator model tends to copy very long sequences of words including whole sentences from the source document (also observed by \citet{Gehrmann2018BottomUpAS}). We use two metrics to evaluate the abstractiveness of the model:
\paragraph{Copy Length: } Table \ref{tab:sequence_length} shows a comparison of the average length (Copy Len) of contiguous copied sequences from the source document (greater than length 3). We observe that the pointer-generator baseline on average copies 16.61 continuous tokens from the source which shows the extractive nature of the model. This indicates that pointer networks, aimed at combining advantages from abstractive and extractive methods by allowing to copy content from the input document, tend to skew towards copying, particularly in this dataset. A consequence of this is that the model fails to interrupt copying at desirable sequence length. In contrast, modeling document structure through \ourmodel reduces the length of copied sequences to 9.13 words on average reducing the bias of copying sentences entirely. This average is closer to the reference (5.07 words) in comparison, without sacrificing task performance. \ourmodel learns to stop when needed, while still generating coherent summaries.

\begin{figure}
    \centering
    \includegraphics[width=0.5\textwidth]{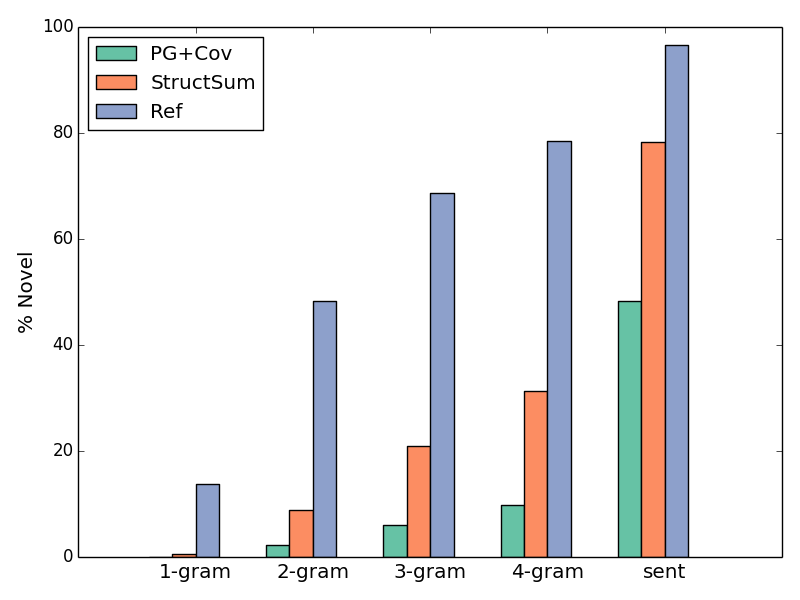}
    \caption{Comparison of \% Novel n-grams between \ourmodel, Pointer-Generator+Coverage and the Reference. Here, ``sent'' indicates full novel sentences.}
    \label{fig:ngrams}
\end{figure}

\paragraph{Novel N-Grams: }The proportion of novel n-grams generated has been used in the literature to measure the degree of abstractiveness of summarization models \cite{see-etal-2017-get}. Figure~\ref{fig:ngrams} compares the percentage of novel n-grams in \ourmodel as compared to the baseline model. Our model produces novel trigrams 21.0\% of the time and copies whole sentences only 21.7\% of the time. In comparison, the pointer-generator network has only 6.1\% novel trigrams and copies entire sentences 51.7\% of the time. This shows that \ourmodel on average generates 14.7\% more novel n-grams in comparison to the pointer-generator baseline.

\subsection{Coverage}
\label{sec:cov}
A direct outcome of copying shorter sequences is being able to cover more content from the source document within given length constraints. We observe that this leads to better summarization performance. We compute coverage by computing the number of source sentences from which contiguous sequences greater than length 3 are copied in the summary. Table \ref{tab:sequence_length} shows a comparison of the coverage of source sentences in the summary content.  While the baseline pointer-generator model only copies from 12.1\% of the source sentences, \ourmodel copies content from 24.0\% of the source sentences. Additionally, the average length of the summaries produced by \ourmodel remains mostly unchanged at 66 words on average compared to 61 of the baseline model. This indicates that \ourmodel produces summaries that draw from a wider selection of sentences from the original article compared to the baseline models.

\begin{figure}
    \centering
    \includegraphics[width=0.5\textwidth]{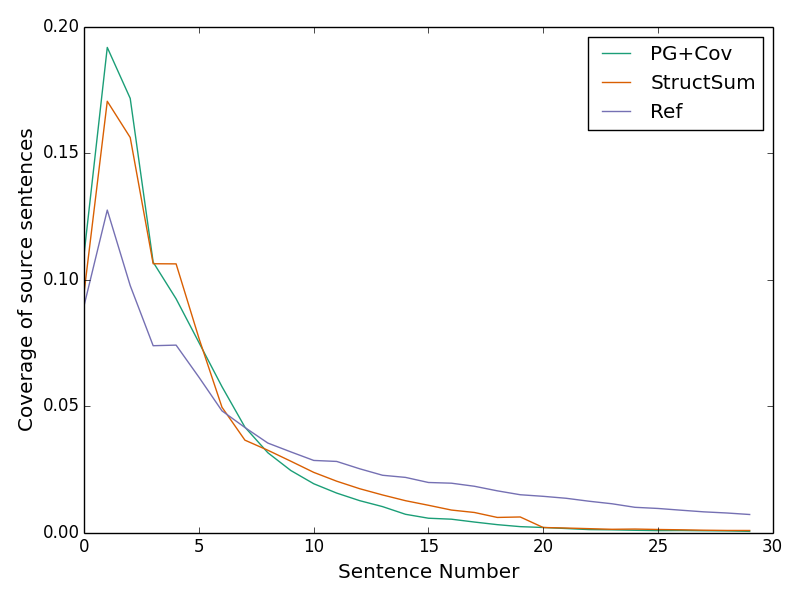}
    \caption{Coverage of source sentences in summary. Here the x-axis is the sentence position in the source article and y-axis shows the normalized count of sentences in that position copied to the summary.}
    \label{fig:coverage}
\end{figure}

\begin{table}[]
    \centering
    \begin{tabular}{@{}lrrrl@{}}
     \toprule
                        & Copy Len & Coverage   \\ \midrule
 PG+Cov            & 16.61 & 12.1 \%    \\ 
     \ourmodel         & 9.13  & 24.0 \%    \\ 
     Reference         & 5.07  & 16.7 \%    \\ 
     \bottomrule
    \end{tabular}
    \caption{Results of analysis of copying and coverage distribution over the source sentences on CNN/DM test set. Copy Len denotes the average length of copied sequences; Coverage -- coverage of source sentences.}
    \label{tab:sequence_length}
\end{table}

\subsection{Layout Bias}
\label{sec:layout}
Neural abstractive summarization methods applied to news articles are typically biased towards selecting and generating summaries based on the first few sentences of the articles. This stems from the structure of news articles, which present the salient information of the article in the first few sentences and expand in the subsequent ones. As a result, the LEAD 3 baseline, which selects the top three sentences of an article, is widely used in the literature as a strong baseline to evaluate summarization models applied to the news domain \citep{narayan-etal-2018-xsum}.  \citet{kryscinski-etal-2019-neural} observed that the current summarization models learn to exploit the layout biases of current datasets and offer limited diversity in their outputs. 

To analyze whether \ourmodel also holds the same layout biases, we compute a distribution of source sentence indices that are used for copying content (copied sequences of length 3 or more are considered). Figure~\ref{fig:coverage} shows the distributions of source sentences covered in the summaries. The coverage of sentences in the reference summaries shows a high proportion of the top 5 sentences of any article being copied to the summary. Additionally, the reference summaries have a smoother tail end distribution with relevant sentences in all positions being copied. It shows that a smooth distribution over all sentences is a desirable feature. We notice that the pointer-generator framework have a stronger bias towards the beginning of the article with a high concentration of copied sentences within the top 5 sentences of the article. In contrast, \ourmodel improves coverage slightly having a lower concentration of top 5 sentences and copies more tail end sentences than the baselines. However, although the modeling of structure does help, our model has a reasonable gap compared to the reference distribution. We see this as an area of improvement and a direction for future work.

\begin{table}[]
    \centering
    \begin{tabular}{@{}llcc@{}}
     \toprule
                    & Coref & NER   & Coref+NER \\
        \midrule
        precision   & 0.29  & 0.19  & 0.33\\
        recall      & 0.11  & 0.08 &  0.09 \\
     \bottomrule    
    \end{tabular}
    \caption{Precision and recall of ES and LS shared edges}
    \label{tab:latent_explicit_distance}
\end{table}

\section{Analysis of Induced Document Structures}

\begin{figure*}
    \centering
    \includegraphics[width=1.0\textwidth]{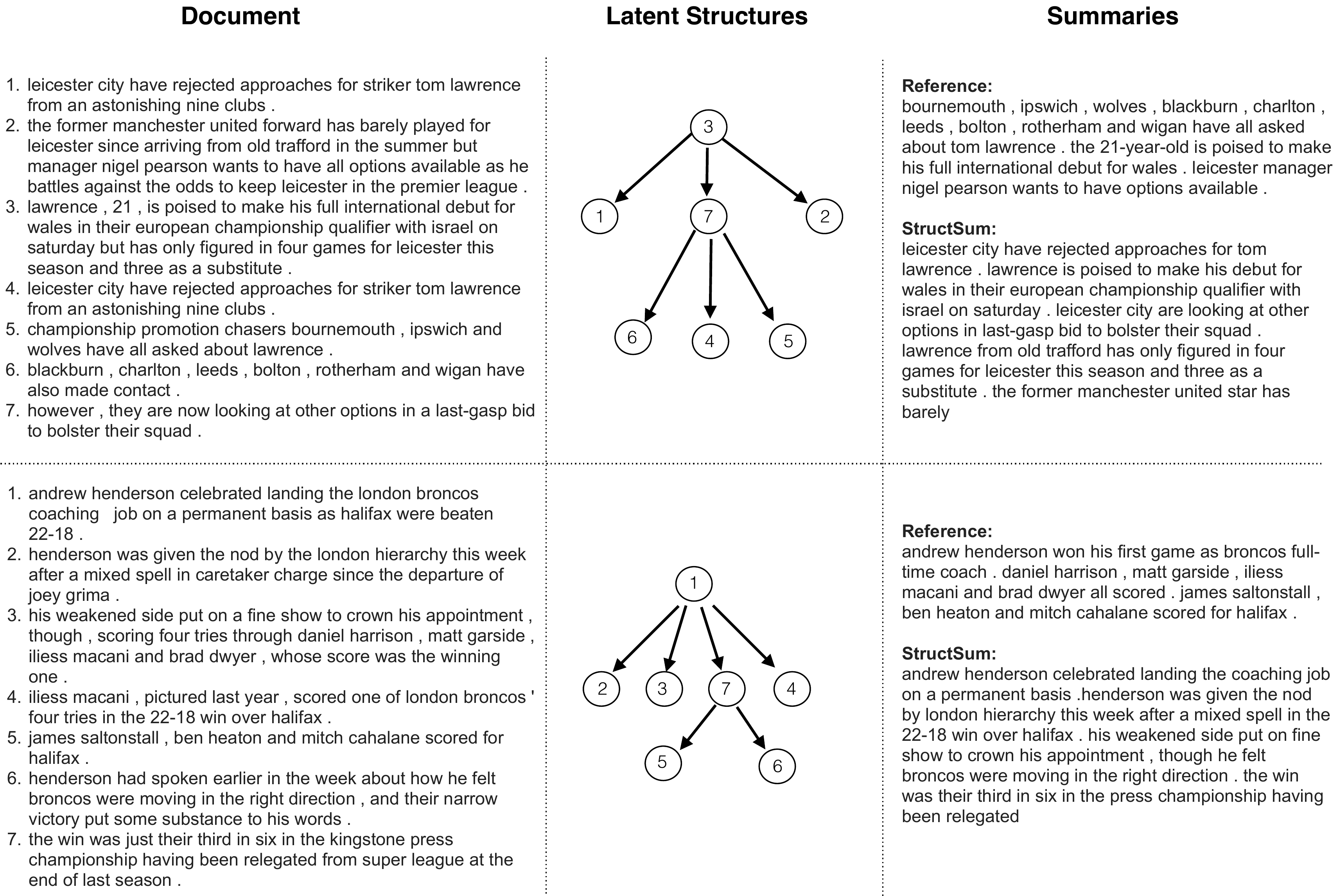}
    \caption{Examples of induced structures and generated summaries.}
    \label{fig:examples}
\end{figure*}

Similar to \citet{Liu2017LearningST}, we also look at the quality of the intermediate structures learned by the model. We use the Chu-Liu-Edmonds algorithm \citep{Chu1965OnTS, edmonds1967optimum} to extract the maximum spanning tree from the attention score matrix as our sentence structure. Table \ref{tab:tree_depth} shows the frequency of various tree depths. We find that the average tree depth is 2.9 and the average proportion of leaf nodes is 88\%, consistent with results from tree induction in document classification \cite{Ferracane2019EvaluatingDI}. Further, we compare latent trees extracted from \ourmodel with undirected graphs based on coreference, on NER, or on both. These are constructed similarly to our explicit coreference based sentence graphs in \Sref{sec:ES} by linking sentences with overlapping coreference mentions or named entities. We measure the similarity between the learned latent trees and the explicit graphs through precision and recall over edges. The results are shown in Table \ref{tab:latent_explicit_distance}. We observe that our latent graphs have low recall with the linguistic graphs showing that our latent graphs do not capture the coreference or named entity overlaps explicitly, suggesting that the latent and explicit structures capture complementary information. 
\ignore{
\begin{table}[]
    \centering
    \begin{tabular}{@{}llll@{}}
     \toprule
        Depth & StructSum \\  
        \midrule
        2 & 29.3\% \\
        3 & 53.7\% \\
        4 & 14.4\% \\
        5+ & 2.6\% \\
     \bottomrule
    \end{tabular}
    \caption{Distribution of latent tree depth.}
    \label{tab:tree_depth}
\end{table}
}
\begin{table}[]
    \centering
    \begin{tabular}{@{}lcccc@{}}
     \toprule
        Depth & 2 & 3 & 4 & 5+ \\
        \midrule
        StructSum  & 29.3\% & 53.7\% & 14.4\% & 2.6\% \\
     \bottomrule
    \end{tabular}
    \caption{Distribution of latent tree depth.}
    \label{tab:tree_depth}
\end{table}

Figure~\ref{fig:examples} shows qualitative examples of induced structures along with summaries from the \ourmodel. The first example shows a tree with sentence 3 chosen as root, which was the key sentence mentioned in the reference. In both examples, the sentences in the lower level of the dependency tree contribute less to the generated summary. Similarly, in the examples source sentences used to generate summaries tend to be closer to the root node. In the first summary, all source content sentences used in the summary are either the root node or within depth 1 of the root node. In the second example, 4 out of 5 source sentences were at depth=1 in the tree. In both examples, generated summaries diverged from the reference by omitting certain sentences used in the reference. These sentences are in the lower section of the tree, providing insights on which sentences were preferred for the summary generation. We also see in example 1 that the latent structures cluster sentences based on the main topic of the document. Sentences 1,2,3 differ from sentences 5,6,7 in the topic discussed and our model clustered the two sets separately. 

\section{Related Work}
Data-driven neural summarization falls into \emph{extractive} \cite{cheng-etal-2016-long, zhang-etal-2018-neural} or \emph{abstractive} \cite{rush-2015-sum, see-etal-2017-get, Gehrmann2018BottomUpAS, chen-bansal-2018-fast}. 
Pointer-generator \citet{see-etal-2017-get} learns to either generate novel in-vocabulary words or copy from the source. It has been the foundation for much work on abstractive summarization \citep{Gehrmann2018BottomUpAS, Hsu2018AUM, Song2018StructureInfusedCM}. Our model extends it by incorporating latent/explicit structure, but these extensions are applicable to any other encoder-decoder architecture. For example, a follow-up study has already shown benefits of our method in multi-document summarization \cite{chowdhury2020neural}.

In pre-neural era, document structure played a critical role in summarization \cite{leskovec2004learning,litvak2008graph,liu-etal-2015-toward,Durrett2016LearningBasedSS,kikuchi2014single}. 
More recently \citet{Song2018StructureInfusedCM} infuse source syntactic structure into the pointer-generator using word-level syntactic features and augmenting them to decoder copy mechanism. In contrast, we model sentence dependencies as latent structures and explicit coreference structures; we do not use heuristics or salient features. \citet{li2018improving} propose structural compression and coverage regularizers incorporating structural bias of target summaries while we model the structure of the source document. \citet{frermann2019inducing} induce latent structures for aspect based summarization, \citet{cohan-etal-2018-discourse} focus on summarization of scientific papers, \citet{Isonuma2019UnsupervisedNS} reviews unsupervised summarization, \citet{mithun-kosseim-2011-discourse} use discourse structures to improve coherence in blog summarization and \citet{ren2018sentence} use sentence relations for multi-document summarization. These are complementary directions to our work. 
To our knowledge, \ourmodel is the first to jointly incorporate latent and explicit document structure in a summarization framework.

\section{Conclusion and Future Work}
In this work, we propose the framework \ourmodel for incorporating latent and explicit document structure in neural abstractive summarization. We introduce a novel explicit-attention module which incorporates external linguistic structures, instantiating it with coreference links. We show that our framework improves the abstractiveness and coverage of generated summaries, and helps mitigate layout biases associated with prior models. We present an extensive evaluation of \ourmodel- along abstractiveness, coverage, and layout quantitatively. 
Future work will investigate the role of document structures in pretrained language models \citep{Lewis2019BARTDS,Liu2019TextSW}.

\section*{Acknowledgements}
The authors are grateful to the anonymous reviewers for their invaluable feedback and to Sandeep Subramanian, Waleed Ammar and Kathryn Mazaitis for their help and support in various stages of the project.
This material is based upon work supported by the DARPA SemaFor and NNSA DOE programs. 
Any opinions, findings, and conclusions or recommendations expressed in this material are those of the author(s) and do not necessarily reflect the views of the DARPA or NNSA.
We would also like to thank Amazon for providing GPU credits.

\bibliography{anthology,eacl2021}
\bibliographystyle{acl_natbib}

\appendix

\end{document}